\documentclass[10pt,twocolumn,letterpaper]{article}

\usepackage{iccv}
\usepackage{times}
\usepackage{epsfig}
\usepackage{graphicx}
\usepackage{amsmath}
\usepackage{amssymb}
\usepackage{amsfonts}
\usepackage{booktabs}
\usepackage{multirow}
\usepackage{verbatim}

\usepackage{color}
\usepackage{subfigure}

\def \H {\mathbf{H}}

\def \E {\mathbf{E}}
\def \RR {\mathbb{R}}

\usepackage{enumerate}

\usepackage[pagebackref=true,breaklinks=true,letterpaper=true,colorlinks,bookmarks=false]{hyperref}

 \iccvfinalcopy 

\def\iccvPaperID{****} 

\ificcvfinal\fi

\begin{document}

\title{Metric Learning for Anti-Compression Facial Forgery Detection}

\author{Shenhao Cao, Qin Zou*, Xiuqing Mao, Zhongyuan Wang\\
School of Computer Science, Wuhan University\\
* {\tt\small qzou@whu.edu.cn}
}

\maketitle
\ificcvfinal\thispagestyle{empty}\fi

\begin{abstract}
  Detecting facial forgery images and videos is an increasingly important topic in multimedia forensics. As forgery images and videos are usually compressed into different formats such as JPEG and H264 when circulating on the Internet, existing forgery-detection methods trained on uncompressed data often suffer from significant performance degradation in identifying them. To solve this problem, we propose a novel anti-compression facial forgery detection framework, which learns a compression-insensitive embedding feature space utilizing both original and compressed forgeries. Specifically, our approach consists of three ideas: (i) extracting compression-insensitive features from both uncompressed and compressed forgeries using an adversarial learning strategy; (ii) learning a robust partition by constructing a metric loss that can reduce the distance of the paired original and compressed images in the embedding space; (iii) improving the accuracy of tampered localization with an attention-transfer module. Experimental results demonstrate that, the proposed method is highly effective in handling both compressed and uncompressed facial forgery images.
\end{abstract}

\section{Introduction}
The rapid development of deep learning, especially generative adversarial networks~\cite{1-DBLP:conf/nips/GoodfellowPMXWOCB14} and variational autoencoders~\cite{2-DBLP:journals/corr/KingmaW13}, enables an attacker to create face forgeries that are indistinguishable by human eyes. Many deep learning-based~\cite{3-DBLP:journals/tog/ThiesZN19,4-DBLP:conf/eccv/WilesKZ18,5-DBLP:conf/iccv/ZakharovSBL19,6-DBLP:conf/cvpr/ChenWYT20} and computer graphics-based~\cite{7-DBLP:journals/tog/Averbuch-ElorCK17,8-DBLP:conf/cvpr/ThiesZSTN16,9-DBLP:journals/tog/ThiesZTSN18} face forgery technologies are proposed. Anyone without professional image editing skills can use customized applications, such as DeepFaceLab~\cite{10-DBLP:journals/corr/abs-2005-05535}, FaceApp and Zao, to create realistic face forgeries. However, those face forgery technologies may be abused without permission, which has great damage to the citizens' portrait and reputation rights and even endangers the national political. Therefore, the face forgery detection technology is particularly important.

\begin{figure}[!t]
	\begin{center}
		\includegraphics[scale=0.35]{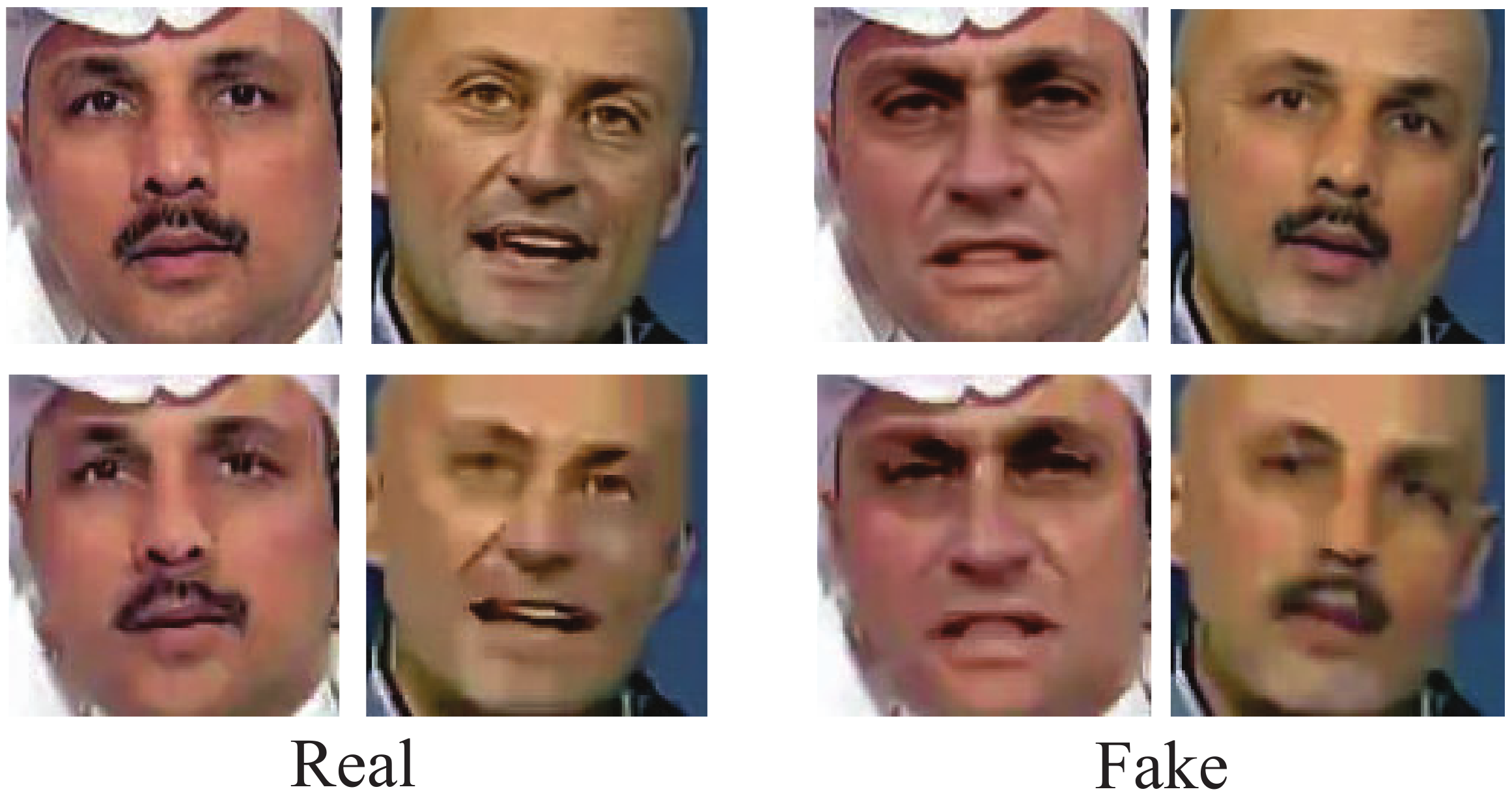}
	\end{center}
	\vspace{-0.1in}
	\caption{Examples of original and compressed facial images. Top row: original high-quality images; Bottom row: the corresponding compressed low-quality images. The left are real facial images, and the right are forgery ones. Compressed forgery images are more difficult to be identified.}
	\label{fig:showdata}
	\vspace{-0.1in}
\end{figure}

Face forgeries, no matter generated by identity swap~\cite{13-DBLP:conf/iccv/SuwajanakornSK15,14-DBLP:conf/iccv/KorshunovaSDT17,15-DBLP:conf/siggraph/NatsumeYM18,16-DBLP:conf/accv/NatsumeYM18,36-DBLP:conf/cvpr/LiBYCW20}, expression swap~\cite{8-DBLP:conf/cvpr/ThiesZSTN16,9-DBLP:journals/tog/ThiesZTSN18,18-DBLP:journals/tog/KimCTXTNPRZT18,19-DBLP:conf/eccv/WuZLQL18,20-DBLP:conf/bmvc/ZhangZHLLL19} or GAN~\cite{21-DBLP:conf/iclr/KarrasALL18,22-DBLP:conf/cvpr/KarrasLA19}, contain forgery artifacts in both color and frequency domains~\cite{23-DBLP:conf/cvpr/WangW0OE20}. Various forgery detection approaches were proposed in the past decade, including the Capsule~\cite{24-DBLP:journals/corr/abs-1910-12467}, FWA~\cite{25-DBLP:conf/cvpr/LiL19c}, HeadPose~\cite{26-DBLP:conf/icassp/YangLL19a}, MesoNet~\cite{27-DBLP:conf/wifs/AfcharNYE18}, Multi-task~\cite{28-DBLP:conf/btas/NguyenFYE19}, Two-stream~\cite{29-DBLP:conf/cvpr/ZhouHMD17}, and Face X-ray~\cite{30-DBLP:conf/cvpr/LiBZYCWG20}, etc. In actual scenarios, media platforms often compress pictures and videos in order to save transmission bandwidth. However, forgery artifacts will fade after compression, which increases the difficulty of face forgery detection. As shown in Fig.~\ref{fig:showdata}, it is difficult to distinguish the authenticity of heavily compressed facial images at the bottom row compared to original images at the top row. XceptionNet~\cite{31-DBLP:conf/iccv/RosslerCVRTN19} achieved 99.26$\%$ accuracy in raw data, but the accuracy decreased to 81.00$\%$ in low quality data. At present, there are still few detection methods specifically for heavily compressed face forgeries. To solve this problem, we propose a two-branch network extracting compression-insensitive features to improve the accuracy of compressed forgery detection.

It is worth noting that, in the quantization process of the compression algorithm, \textit{e.g.} JPEG, the high frequency and partial mid frequency DCT coefficients are quantized to 0, in order to reduce the code volume. That is to say, high frequency features and partial mid frequency features of the forgery artifacts are lost after compression. However, the original image contains the complete forgery information in all three frequency bands of low frequency, mid frequency and high frequency. Besides, the compression algorithm does not modify the manipulation region so that paired different compression level images share a common manipulated region. Therefore, we make the assumption that the forgery information of the compressed image are a subset of the forgery information of the original image. Existing data-driven methods training a two-classification network~\cite{24-DBLP:journals/corr/abs-1910-12467,27-DBLP:conf/wifs/AfcharNYE18,29-DBLP:conf/cvpr/ZhouHMD17,30-DBLP:conf/cvpr/LiBZYCWG20}, or using multi-task joint training~\cite{28-DBLP:conf/btas/NguyenFYE19} cannot completely extract the effective forgery information of compressed images. Our idea is to use the containment relationship between the forgery information in the compressed image and the original image, so that the network extracts the intersection of the two kinds of forgery information, which is fully forgery information in the compressed image.

We propose a two-branch network taking the paired images from different compression levels as the input, and combine the adversarial learning and metric learning~\cite{33-DBLP:conf/eccv/WenZL016, 34-DBLP:conf/eccv/MasiKMGA20} to train the network. The network learns the common feature of the two images in a compression-insensitive embedding feature space. Although there is a strong correlation between the paired compressed image and the original image, the use of different network branches will map the images of two compression levels to different feature spaces. Inspired by the idea of cross-modal retrieval~\cite{32-DBLP:conf/cvpr/WangSLLH19}, we use a discriminator to discriminate whether the feature comes from the compressed image or the original image. When it is well trained, the final discriminator cannot distinguish the source of the feature. That is to say, the compressed image and the original image are mapped to a common feature space with the anti-compression characteristics. Besides, distance between the paired compression and the original images in the feature space should be reduced. To achieve it, we introduce the metric-learning strategy to reduce the feature distance of compressed and uncompressed forgeries. In addition, we introduce an attention module on both branches to push the network concentrating on the tampered region, and transfer attention information to the low-quality branch to improve the tampered- or manipulated-region prediction.

Overall, in this paper we make the following contributions:
\begin{enumerate}[  $\vcenter{\hbox{\small$\bullet$}}$]
\item A two-branch network is proposed which maps forgery artifacts from paired original and compressed forgeries to a compression-insensitive embedding feature space by using an adversarial-learning strategy;	
\item A metric loss function is introduced that reduces the distance of the paired images of two different compression levels in the embedding feature space, which further encourages the network to extract the compression-insensitive features;
\item An attention transfer module is proposed, which transfers information from the high-quality branch to the low-quality branch, and improves the accuracy of manipulated-region prediction on the low-quality branch.
\end{enumerate}

\begin{figure*}
	\begin{center}
		\includegraphics[scale=0.13]{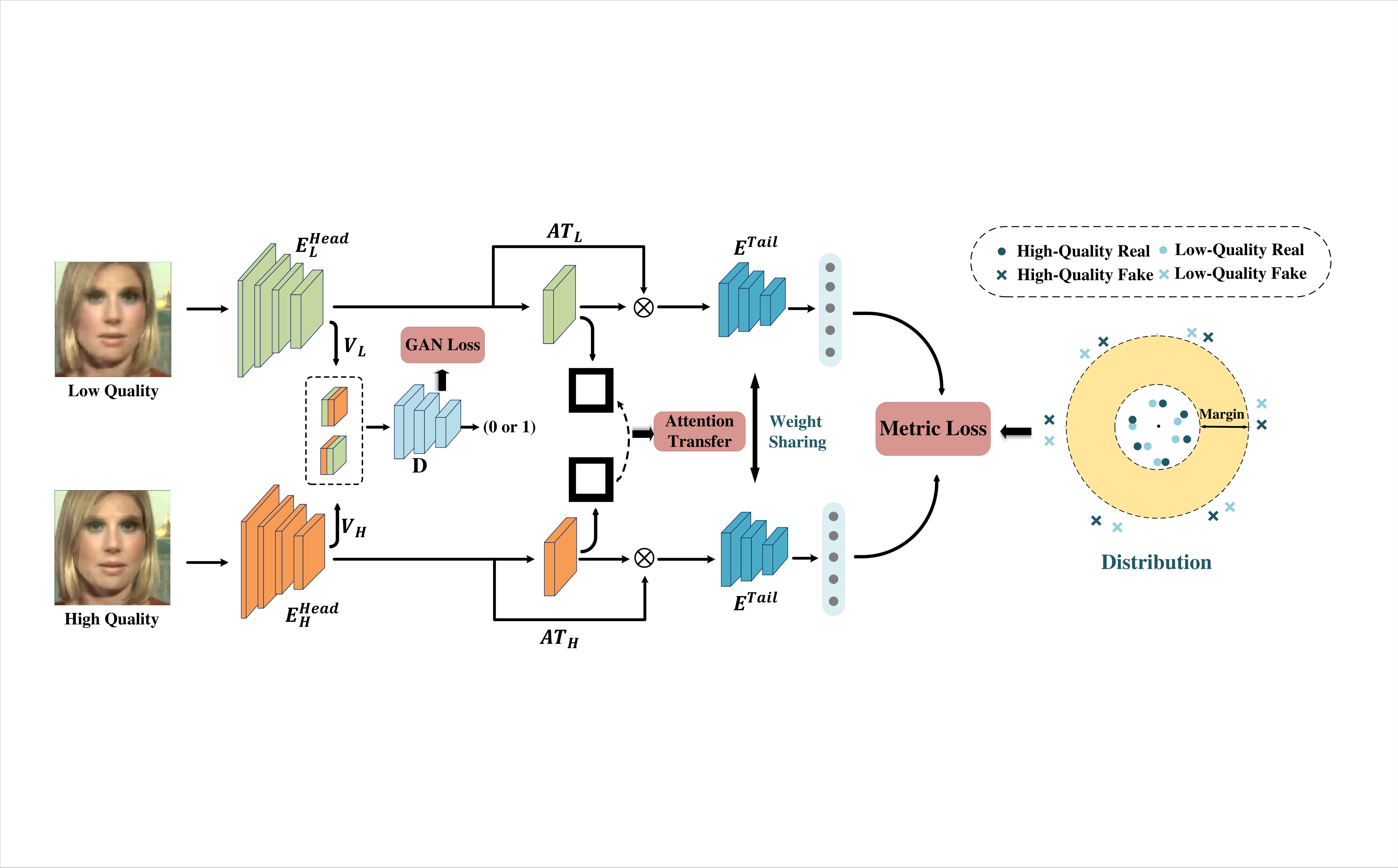}
	\end{center}
	\vspace{-0.1in}
	\caption{The architecture of our two-branch network. Each branch consists of a head encoder, an attention layer, and a weight-shared tail encoder. In the training phase, paired high-quality and low-quality images are separately fed into the bottom and top branches. The top branch for low-quality is used in the test phase. }
	\label{fig:pipeline}
	\vspace{-0.1in}
\end{figure*}

\section{Related Work}
In this section, we briefly overview the work on facial forgery and forgery detection.
\subsection{Facial Forgery Technologies}
Face forgery technologies can be mainly divided into three categories: GAN-based face synthesis, face swap and face reenactment. Recent improvements in Generative Adversarial Networks~\cite{21-DBLP:conf/iclr/KarrasALL18,22-DBLP:conf/cvpr/KarrasLA19,35-DBLP:conf/cvpr/KarrasLAHLA20} enable the generation of an entire high-resolution face image with few visible artifacts. ProGAN~\cite{21-DBLP:conf/iclr/KarrasALL18} generates images layer by layer from coarse to fine, and can generate extremely realistic high-definition face images. StyleGAN~\cite{22-DBLP:conf/cvpr/KarrasLA19} decouples the input vector, Enable the generator to generate face images with specified characteristics, such as hair color, skin tone, etc. StyleGAN2~\cite{35-DBLP:conf/cvpr/KarrasLAHLA20} reduces visual artifacts and further improves the image quality.

Face swap,well known as Deepfakes, replaces a face in images or videos with another and face reenactment transfers expressions from one person to another. Graphics-based approaches, especially 3D model technology, are widely used for face swap~\cite{13-DBLP:conf/iccv/SuwajanakornSK15} and face reenactment~\cite{8-DBLP:conf/cvpr/ThiesZSTN16,18-DBLP:journals/tog/KimCTXTNPRZT18}. Suwajanakorn et al.~\cite{13-DBLP:conf/iccv/SuwajanakornSK15} uses the 3D model method for face swap, and achieved excellent results using hundreds of original face and target face images. Face2face~\cite{8-DBLP:conf/cvpr/ThiesZSTN16} proposes a real-time face reenactment framework for RGB videos. The extended work~\cite{18-DBLP:journals/tog/KimCTXTNPRZT18} proposed a method that can transfer expression, 3D head pose, and eye blinking among videos.

Deep learning methods~\cite{15-DBLP:conf/siggraph/NatsumeYM18,16-DBLP:conf/accv/NatsumeYM18,19-DBLP:conf/eccv/WuZLQL18,20-DBLP:conf/bmvc/ZhangZHLLL19,36-DBLP:conf/cvpr/LiBYCW20} are more effective in synthesizing or manipulating faces. RSGAN~\cite{15-DBLP:conf/siggraph/NatsumeYM18} and FSGAN~\cite{16-DBLP:conf/accv/NatsumeYM18} combine Variational Autoencoder and GAN, encoding face area and non-face area separately and reconstruct an entire face swapped image. FaceShifter~\cite{36-DBLP:conf/cvpr/LiBYCW20} uses only a few images to generate high fidelity face swapped images with facial occlusions. As for face reenactment, ReenactGAN~\cite{19-DBLP:conf/eccv/WuZLQL18} extracts face contour of source image, transfers contour to the target and reconstruct the image only using a feedforward network structure. Zhang et al.~\cite{20-DBLP:conf/bmvc/ZhangZHLLL19} decomposed the face image into appearance space and shape space, using only one source and one target image to achieve many-to-many expression transfer, and can generate more realistic beards and hairs.

\subsection{Forgery Detection Technologies}
Current detection methods generally regard the detection problem as a binary classification problem. Statistics-based features and neural networks are popular for GAN generated images detection. Difference in color distribution~\cite{39-DBLP:journals/corr/abs-1812-08247} and texture distribution~\cite{40-DBLP:conf/cvpr/LiuQT20} between GAN generated images and real images are utilized for classification. Gaussian noise preprocessing~\cite{41-DBLP:conf/ccbr/Xuan0WD19} is proved to improve the representation and generalization capabilities of neural networks. FakeSpotter~\cite{38-DBLP:conf/ijcai/WangJMXHWL20}monitors the activation value of neurons for filtering effective features to train a binary classifier.

Various methods are proposed for detecting forgeries generated by face swap and face reenactment technologies. Visual aircrafts in the eyes and teeth area~\cite{43-DBLP:conf/wacv/MaternRS19} and 3D head pose~\cite{26-DBLP:conf/icassp/YangLL19a} are also used. Deep learning based methods are used to automatically extract discriminative features. Li et al.~\cite{25-DBLP:conf/cvpr/LiL19c} uses CNN to capture the aircrafts introduced by splicing process to distinguish Deepfake videos. In ~\cite{31-DBLP:conf/iccv/RosslerCVRTN19} XceptionNet~\cite{44-DBLP:conf/cvpr/Chollet17} is directly trained as a binary classifier on FaceForensics++~\cite{31-DBLP:conf/iccv/RosslerCVRTN19}. Besides classification, many methods focus on localizing the tampered regions. Muti-task learning strategy~\cite{28-DBLP:conf/btas/NguyenFYE19,45-DBLP:journals/jvcir/SalloumRK18,46-DBLP:journals/corr/abs-1910-05455} by joint training a classification and segmentation network can simultaneously detect the tampered images and locate the tampered region. Stehouwer et al.~\cite{47-DBLP:conf/cvpr/DangLS0020} present a localization architecture through an attention mechanism where attention map denotes the interested manipulation region.

State-of-art methods trained on high quality dataset are not suitable for detecting compressed manipulations, while discriminative features fade after compression. Previous studies~\cite{31-DBLP:conf/iccv/RosslerCVRTN19,48-DBLP:journals/corr/abs-1909-12962} verify that accuracy drops significantly tested on compressed manipulations. However, there are still few detection methods~\cite{49-DBLP:conf/eccv/QianYSCS20,50-DBLP:conf/iwbf/KumarBV20} specifically for compressed face forgeries. F3-Net~\cite{49-DBLP:conf/eccv/QianYSCS20} introduces frequency into the face forgery detection, taking advantages of two different but complementary frequency-aware clues and improving performance on low quality media. Kumar et al.~\cite{50-DBLP:conf/iwbf/KumarBV20} applies a triplet loss to achieve higher accuracy on compressed videos.

\section{Methods}
\subsection{Motivation}
We make an assumption that the discriminative features of low-quality forged images are a subset of high-quality forged images, mainly based on the following two points: (i) Quantification. In the image compression algorithm, an image is converted to the DCT domain in units of blocks. In the next quantization operation, part of the DCT coefficients representing high frequency information are set to 0 according to the quantization table of different compression levels, while the low-frequency and mid-frequency coefficients mostly remain unchanged. Therefore, only high-frequency discriminative features are lost. (ii) Manipulated Region. The compression algorithm does not modify the manipulated region so that features of paired different compression level images representing localization should be similar.

Based on this containment relationship, an embedding function can be learned to extract intersection of paired features, which is retained after compression and can be also called as compression-robust features. The most intuitive idea is to reduce the feature distance between the paired high-quality and low-quality images. A pairwise cosine loss or L2 loss can be used to learn this embedding. However, these approaches suffer from several serious challenges:(i) Compared with original images, compressed images have poorer quality and lower resolution. That's to say, the distributions are different and images with different compression levels will be mapped to different feature spaces. Our approach try to align these feature spaces. (ii) The shared manipulation regions are not concentrated by embedding functions, as a result, information representing localization can be lost. (iii) Directly reducing the encoding distance of the feature is not enough to make the feature more discriminative.

To solve these problems, we propose a novel end-to-end framework consisting of two branches for learning compression-robust embeddings. During the training phase, the two branches take paired high-quality and low-quality images as input respectively while only low-quality branch is used in testing process. To address distribution alignment, adversarial training strategy is used to ensure features across different compression levels following the same distribution. To address the second localization problem, we use attention mechanism to highlight the manipulated image regions and transfer attention map from the high-quality branch to low-quality, which enforces two branches concentrate on the same region. Finally, we use a metric loss that compresses the realistic faces and pushes away the fake faces in the feature space and further add a L2-loss to reduce distances across compression levels.

\subsection{Architecture}
More formally, given a set of different compression level pairs $(\mathbf{h}^t, \mathbf{l}^t)$ for $t = 1, \dots, T$, where a high-quality image $\mathbf{h}^t \in \mathbf{H}$ and a low-quality $ \mathbf{l}^t \in \mathbf{L}$ ($\mathbf{H}$ and $\mathbf{L}$ correspond to the high-quality and low quality images respectively), our goal is to learn embedding functions $\E_H: \H \rightarrow \RR^d$ and $\E_L: \mathbf{L} \rightarrow \RR^d$ which encode two paired images into  $d-$dimensional  vectors, respectively. In the process of the feed-forward pipeline, the high-quality images  $\mathbf{h}$ and low-quality images $\mathbf{l}$ are fed into two CNNs, where the tail sub-networks are weight shared. The high-quality branch $\E_H$ can be divided into three part: head sub-network $\E^{Head}_H$, attention layer $\mathbf{AT}_H$ and tail sub-network $\E^{Tail}$. Identically, $\E_L$ consists of  $\E^{Head}_L$, $\mathbf{AT}_L$ and $\E^{Tail}$. The two branches share the common tail sub-network $\E^{Tail}$. The head networks $\E^{Head}_H$ and $\E^{Head}_L$ give us mid-level features $V_H$ and $V_L$. Aimed at aligning feature spaces across compression levels, $V_H$ and $V_L$ are fed into a discriminator $\mathbf{D}$.

These mid-level features are also fed into attention layer generating corresponding attention map $M_H$ and $M_L$. Attention map is used to highlight the manipulated regions so that $M_H$ should be close to $M_L$. Considering the high-quality branch has more information than the low-quality branch, we transfer attention information from $M_H$ to $M_L$. In the next step, $V_H$ and $V_L$ are modified based on activation value of attention maps, as the input of the tail sub-network. Finally, the high-quality branch gives us high-level features  $C_h \in \RR^d$, and the low-quality branch gives us high-level features  $C_l \in \RR^d$. The embedding functions should ensure that $C_h$ is close to $C_h$ in the feature space.

This framework is trained under three objectives: to obtain a feature space that aligns distributions across compression levels; to transfer attention information from high-quality branch to low-quality branch making two branches share manipulated regions; and to reduce feature distances in the embedding space where the realistic faces are close to a fixed point and fake faces are far away from that point. The total objective of our framework is given as:
\begin{equation}
\begin{aligned}
\label{eq:main}
\mathcal{L} =  \mathcal{L}_{Dis} + \lambda_1 \mathcal{L}_{Gan} + \lambda_2 \mathcal{L}_{AT},
\end{aligned}
\end{equation}
\label{total_formulation}
where $\lambda_1$ and $\lambda_2$  are two weight parameters. The distribution alignment component $\mathcal{L}_{Gan} (V_H, V_L)$ using an adversarial loss to align the two distributions, operating on mid-level features $V_H$ and $V_L$. The attention transfer component $\mathcal{L}_{AT}$ calculates transfer loss between attention map $M_H$ and $M_L$, and also losses between ground truth and attention maps. The reducing coding distance component $\mathcal{L}_{Dis}$ computes metric losses of $C_h$ and $C_l$, and also L2-loss between $C_h$ and $C_l$.

\subsection{Adversarial Learning for Distributions Alignment}
Distributions of the encoded features from high-quality and low-quality images can be very different, resulting in slower convergence. Aligning distributions of features is effective for optimization. Inspired by modality alignment process in cross-modal retrieval task\cite{32-DBLP:conf/cvpr/WangSLLH19}, we use an adversarial loss to align the distributions of the mid-level features $V_H$ and $V_L$. Different from \cite{32-DBLP:conf/cvpr/WangSLLH19} using a discriminator to directly identify which modality the feature belongs to, we can't distinguish whether the feature $V_H$ and $V_L$ come from high-quality images or low-quality images. It is worth noting that an image after heavy compression may have higher quality than another image with light compression because samples in public datasets may have the same quality label even they have different qualities. Therefore, $\mathbf{l}^{t1}$ may have higher quality than $\mathbf{h}^{t2}$ where $t1 \neq t2$, and  $\mathbf{l}^{t1}$ definitely has a lower quality than $\mathbf{h}^{t2}$ where $t1 = t2$.

We have paired feature maps $V_H = \E^{Head}_H(\mathbf{h})\in\mathbb{R}^{H\times W\times C}$ and $V_L = \E^{Head}_L(\mathbf{l})\in\mathbb{R}^{H\times W\times C}$, where $H$, $W$, $C$ are height, width, and the number of channels, respectively. We concatenate $V_H$ and $V_L$ along channel and take the concatenated feature maps $concat(V_H, V_L)\in\mathbb{R}^{H\times W\times 2C}$ and $concat(V_L,V_H)\in\mathbb{R}^{H\times W\times 2C}$ as inputs of a discriminator $\mathbf{D}$. $\mathbf{D}$ trys to distinguish weather the concatenated feature comes from $concat(V_H, V_L)$ or $concat(V_L, V_H)$. We achieve a common feature representation so that $\mathbf{D}$ cannot identify the combination order of the concatenated feature. WGAN-GP is adopted in our experiment empirically. The objective is given as:
\begin{equation}
\begin{aligned}
\mathcal{L}_{Gan} =  & \mathbb{E}_{(V_H, V_L)} [\log D(\E^{Head}_H(concat(V_H, V_L))]  + \\
& \mathbb{E}_{(V_L, V_H)} [\log(1 - D(\E^{Head}_L(concat(V_L, V_H)))],
\end{aligned}
\end{equation}
and is solved by a min-max optimization:
$$ \min_{\E^{Head}_H, \E^{Head}_L} \max_{D} \mathcal{L}_{Gan}. $$

\subsection{Attention Transfer for Manipulated- Region Prediction}
As shown in Fig.~\ref{fig:pipeline}, the inputs of attention layer are feature maps $V_H\in\mathbb{R}^{H\times W\times C}$ and $V_L\in\mathbb{R}^{H\times W\times C}$. Then, we can generate attention maps $M_H=\mathbf{AT}_H(V_H)\in\mathbb{R}^{H\times W}$ and $M_L=\mathbf{AT}_L(V_L)\in\mathbb{R}^{H\times W}$. We use a direct regression method~\cite{47-DBLP:conf/cvpr/DangLS0020} as the attention layer, which is a convolution layer follow by a  $\mathbf{Sigmoid}$ function. The pixel value of the attention map is close to 0 for the real regions and 1 for the fake regions. The output of the attention is refined by attention map to help the following network concentrate on fake regions. The output feature maps ${V_H}'$ and ${V_L}'$ are calculated as:
\begin{equation}
{V_H}' = V_H \odot M_H,
\end{equation}\label{eq:attention1}\vspace{-3mm}
\begin{equation}
{V_L}' = V_L \odot M_L,
\end{equation}\label{eq:attention2}
where $\odot$ denotes element-wise multiplication.

The pixel-level intensity indicates the probability whether the corresponding region comes from a fake image. We treat each pixel of an attention map as a binary classification problem and cross-entropy loss is used for each pixel as attention loss. Considering that the pixel-level intensity of two attention maps should be closer, we transfer attention information from $M_H$ to $M_L$ using L2-loss, inspired by\cite{53-DBLP:conf/iclr/ZagoruykoK17}. The total attention transfer loss is calculated as
\begin{equation}
\begin{aligned}
\mathcal{L}_{AT} =
& \frac{1}{N} \sum || M_{gt}log(M_H) + (1 -M_{gt})log(1-log(M_H))   ||_1 + \\
& \frac{1}{N} \sum || M_{gt}log(M_L) + (1 -M_{gt})log(1-log(M_L))   ||_1 + \\
& ||\frac{M_H}{||{M_H}||_2}-{\frac{M_L}{||{M_L}||_2}} ||_2,
\label{eq:attention_transfer_loss}
\end{aligned}
\end{equation}
where ${M}_{gt}$ is the ground truth manipulation mask, and $N$ is the number of pixels of an attention map. We use all zeros as the ${M}_{gt}$ for real faces.

\subsection{Metric Loss to Reduce the Feature Distance}
After high-quality and low-quality images are passed through the encoder networks, high-level features $C_h \in \RR^d$ and $C_l \in \RR^d$ are obtained. Our goal is to minimize the distance between $C_h$ and $C_l$ in the feature space. In \cite{29-DBLP:conf/cvpr/ZhouHMD17}, it proposes a metric loss to compresses the realistic faces and pushes away the fake faces in the feature space. We find that this loss is fast to converge cooperated with attention layer. More specifically, taking a realistic image as input, pixel intensity in attention map is close to 0 so that the activation is weakened by attention layer. On the contrary, high activation is obtained for a fake image. We directly pull realistic faces close to the origin and push fake faces away from the origin in the embedding space. The objective $\mathcal{L}_{Dis}$ is formulated as:
\begin{equation}
\begin{aligned}
\mathcal{L}_{Dis} =
& \sum_{{\mathbf{h}^t} \in {Real}} \max(0,||{C_h}||_2-r^-)~+ \\
& \sum_{{\mathbf{h}^t} \in {Fake}} \max(0,r^+ -||{C_h}||_2)~+ \\
& \sum_{{\mathbf{l}^t} \in {Real}} \max(0,||{C_l}||_2-r^-)~+ \\
& \sum_{{\mathbf{l}^t} \in {Fake}} \max(0,r^+ -||{C_l}||_2)~+\\
& \lambda_3\sum_{({\mathbf{l}^t}, {\mathbf{h}^t})} ||{C_h} - {C_l}||_2.
\label{eq:loss}
\end{aligned}
\end{equation}
where $\lambda_3$ is a trade-off parameter.

\section{Experiments}
\subsection{Dataset and Evaluation Metrics}

\begin{figure*}[!t]
	\begin{center}
		\includegraphics[width=0.76\linewidth]{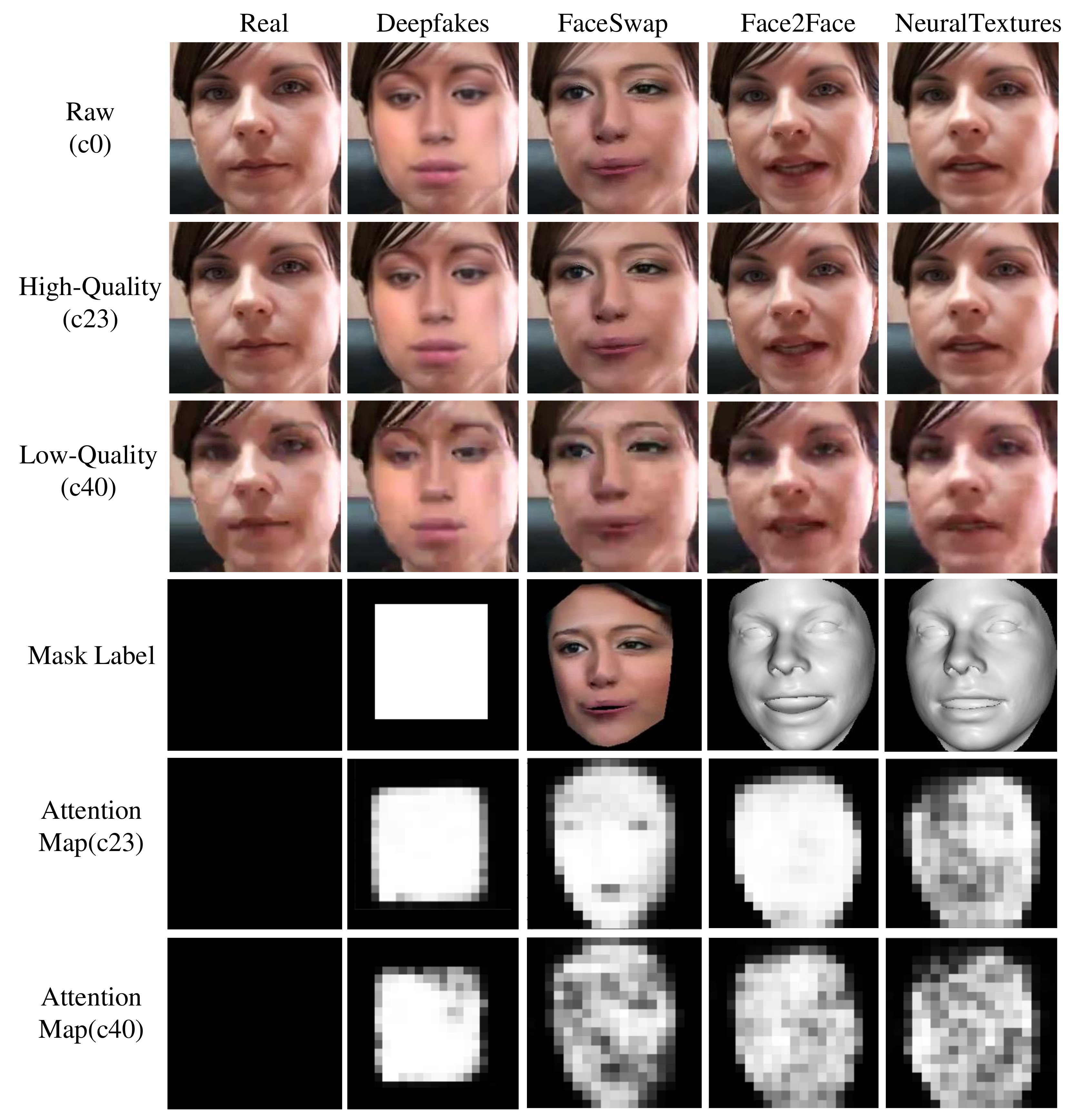}
	\end{center}
	\vspace{-0.1in}
	\caption{Attention maps predicted on a sample facial image of Forenisics++ dataset. }
	\label{fig:mask}
	\vspace{-0.1in}
\end{figure*}

We conduct ablation studies and comparisons on the challenging FaceForensics++\cite{31-DBLP:conf/iccv/RosslerCVRTN19} dataset.  FaceForensics++ dataset contains 1000 real videos, and each real video corresponds to four manipulated types of fake videos. Therefore there are 5000 videos in total. Each video has three compression levels, i.e., RAW, High Quality (HQ) and Low Quality (LQ), which is well tailored for our task. Besides, each fake video corresponds to a mask video which highlights the manipulated region. Following setting in \cite{31-DBLP:conf/iccv/RosslerCVRTN19}, we use 720 real videos and corresponding fake videos for training, 140 videos for validation and 140 videos for testing. We sample 270 frames each video and use the method proposed in \cite{8-DBLP:conf/cvpr/ThiesZSTN16} to obtain the face region of the frame and crop the face region enlarged by a factor of 1.3. The enlarged face region of a fake frame is also used to crop the corresponding mask frame to make ground truth for attention map. As shown in Fig.~\ref{fig:mask}, the first four rows display the examples of three different compression-level images and the manipulated masks of the FaceForensics++ dataset.

Following~\cite{31-DBLP:conf/iccv/RosslerCVRTN19,49-DBLP:conf/eccv/QianYSCS20}, the accuracy score (ACC) is employed as an evaluation metric in our experiments. Following \cite{49-DBLP:conf/eccv/QianYSCS20,30-DBLP:conf/cvpr/LiBZYCWG20,47-DBLP:conf/cvpr/DangLS0020,cao2011cvpr}, we also employ the Area Under the Receiver Operating Characteristic Curve (AUC) as an evaluation metric for classification. Besides, considering that the True Accept Rate (TAR) at a low False Accept Rate (FAR) is widely used for a classification system\cite{49-DBLP:conf/eccv/QianYSCS20, 47-DBLP:conf/cvpr/DangLS0020}, we also report TAR at FAR of $0.01\%$ (denoted as TAR$_{0.01\%}$) and TAR at FAR of $0.1\%$ (denoted as TAR$_{0.1\%}$) as the classification evaluation metrics. Following~\cite{47-DBLP:conf/cvpr/DangLS0020}, to evaluate the localization accuracy of attention maps, the Pixel-wish Binary Classification Accuracy~(PBCA) is used, which measures the classification accuracy by treating each pixel as an independent sample. The attention map is transformed to a binary map at a threshold of 0.5, and PBCA is calculated based on the binary map and the ground-truth mask.

\subsection{Implementation Details}
To obtain the ground truth manipulation mask of a fake image, the cropped mask frame is transformed into grayscale, divided by 255 and converted to a binary map at threshold of 0.1.
In our experiments, Trade-off parameter $\lambda_1$ is set to 0.001, $\lambda_2$ is set to 1  and $\lambda_3$ is set to 0.1. We use Adam optimizer at a learning rate of 0.0001 and a batchsize of 32. The training phase is stopped when the loss doesn't reduce on validation dataset. We augment the size of real images to balance the number of fake images and real images. We use XceptionNet~\cite{44-DBLP:conf/cvpr/Chollet17} pre-trained on ImageNet as backbone networks for both two branches. The newly introduced attention layer and discriminator are randomly initialized. The attention layer is inserted between Block 7 and Block 8 of the middle flow and feature maps of Block 7 are fed to the discriminator for both two branches. To fuse the concatenated features, we use 1$\times$1 convolution kernel for the first convolutional layer of the discriminator. Given that $d = 2048$, $r^-$ is set to $0.1$ and $r^+$ is set to $18.0$ for metric learning. In the test phase, an image is classified as fake if the L2 distance between the origin and the embedding feature is larger than $\frac{r^- + r^+}{2}$.

\begin{table}[!t]
	\begin{center}
		\fontsize{8pt}{12pt}\selectfont
		\caption{Performance of different methods on FaceForenisics++(LQ). Some of results are from \cite{49-DBLP:conf/eccv/QianYSCS20}.}
		\resizebox{0.75\linewidth}{!}{
			\begin{tabular}{l|c|c}
				\toprule
				Methods & ACC & AUC \\
				\midrule
				Steganalysis~\cite{51-DBLP:journals/tifs/FridrichK12}   & $55.98$  & $-$ \\
				CustomPooling CNN~\cite{52-DBLP:conf/wifs/RahmouniNYE17}  & $61.18$  & $-$ \\
				MesoNet~\cite{27-DBLP:conf/wifs/AfcharNYE18}  & $70.47$  & $-$ \\
				Face X-ray~\cite{30-DBLP:conf/cvpr/LiBZYCWG20}  & $-$  & $61.60$ \\
				XceptionNet~\cite{44-DBLP:conf/cvpr/Chollet17} & $86.52$  & $90.32$ \\
				Ours & $87.69$ & $91.54$ \\
				\bottomrule
			\end{tabular}
		}
		
		\label{tab1:results}
	\end{center}
	\vspace{-4mm}
\end{table}

\begin{table}[!t]
	\begin{center}
		\fontsize{8pt}{12pt}\selectfont
		\caption{Accuracy of manipulation-specific forgery detectors on the FaceForenisics++(LQ) datasets. The names of the datasets are abbreviated as DF: DeepFakes, F2F: Face2Face, FS: FaceSwap, and NT: NeuralTextures. Some of results are from~\cite{44-DBLP:conf/cvpr/Chollet17}.}
		\begin{tabular}{l|c|c|c|c}
			\toprule
			Methods & DF & F2F & FS & NT \\
			\midrule
			Steganalysis~\cite{51-DBLP:journals/tifs/FridrichK12} & 65.58 & 57.55 & 60.58 & 60.69\\
			Cozzolino et al.\cite{54-DBLP:conf/ih/CozzolinoPV17}    & 68.26 & 59.38 & 62.08 & 62.42\\
			Bayar and Stamm\cite{55-DBLP:conf/ih/BayarS16}    & 80.95 & 77.30 & 76.83 & 72.38\\
			CustomPooling CNN\cite{52-DBLP:conf/wifs/RahmouniNYE17}     & 73.25 & 62.33 & 67.08 & 62.59\\
			MesoNet \cite{27-DBLP:conf/wifs/AfcharNYE18}            & 89.52 & 84.44 & 83.56 & 75.74\\
			XceptionNet \cite{44-DBLP:conf/cvpr/Chollet17}        & 94.36 & 90.27 & 93.25 & 79.53\\
			Ours               & 95.38 & 91.46 & 94.18 & 81.07\\
			\bottomrule
		\end{tabular}
		\label{tab:all-method}
	\end{center}
	\vspace{-4mm}
\end{table}

\subsection{Comparisons with Other Methods}
We compare our method with other previous face forgery detection methods on the FaceForensics++(LQ) dataset. Table~\ref{tab1:results} shows the AUC and ACC obtained by various methods trained and tested on the whole FaceForenisics++(LQ) dataset. We re-run the XceptionNet~\cite{44-DBLP:conf/cvpr/Chollet17} on our dataset partition and other results are from \cite{49-DBLP:conf/eccv/QianYSCS20}. The comparison is not too strict due to the difference of dataset partition. Table~\ref{tab:all-method} shows the accuracies of the manipulation-specific forgery detectors.

\begin{figure}[!t]
	\begin{center}
		\includegraphics[width=0.99\columnwidth]{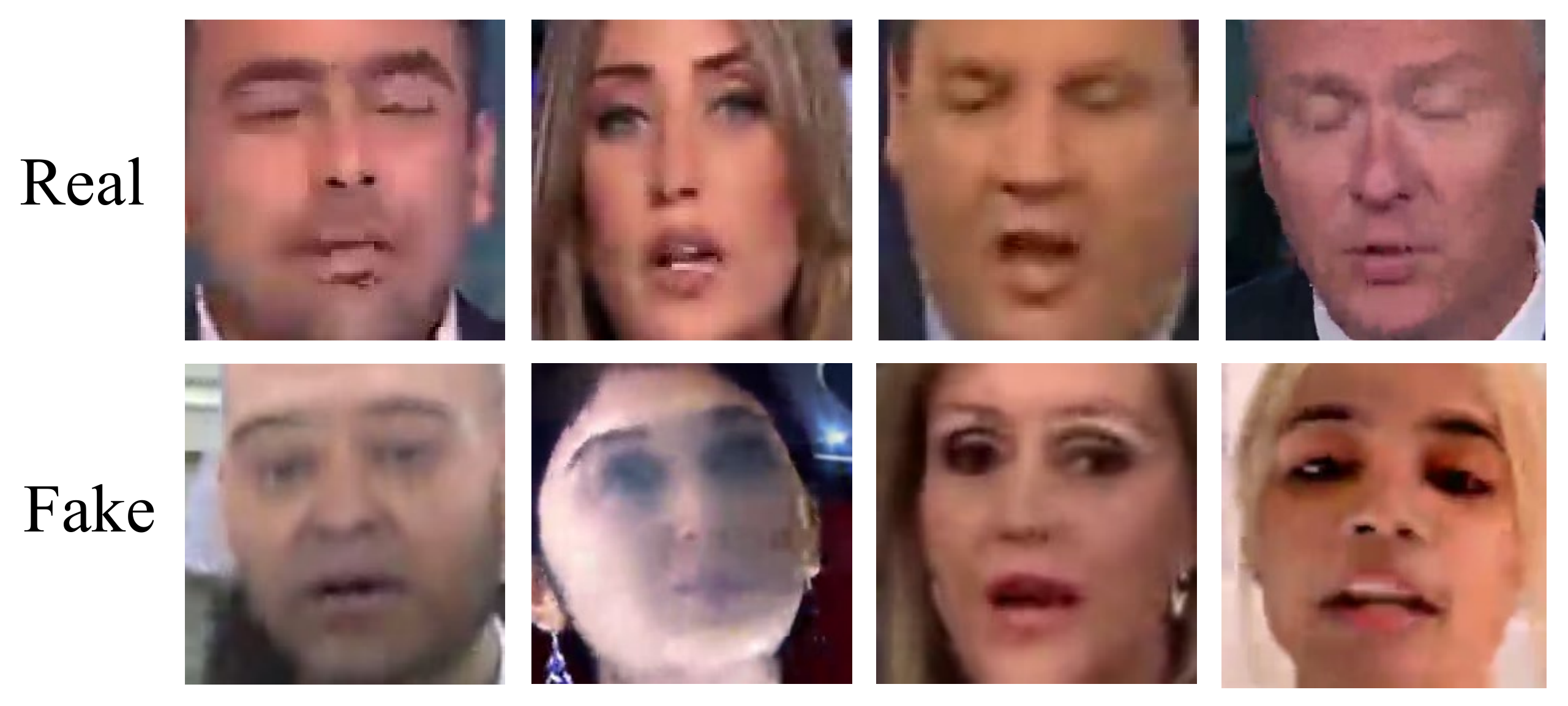}
	\end{center}
	\vspace{-0.1in}
	\caption{ Sample facial images that are correctly classified by our method while falsely classified by XceptionNet.}
	\label{fig:better_results}
	\vspace{-0.1in}
\end{figure}

The results in Table~\ref{tab1:results} and Table~\ref{tab:all-method} show that, our method outperforms the Xception baseline and other previous methods. Compared with XceptionNet, our method achieves an improvement of about $1.17\%$ in ACC and $1.22\%$ in AUC on the whole FaceForensics++(LQ) dataset. When evaluated on the manipulation-specific dataset, our method obtains a promotion on accuracy of about $1.0\%$ , $1.2\%$, $1.1\%$, $1.5\%$ for Deepfakes, Face2Face, FaceSwap and NeuralTextures, respectively. One possible reason for this result is that, the proposed method benefits from the proposed metric-learning strategy, attention transfer, and the adversarial learning, which helps in learning a compression-insensitive feature embedding space using paired high-quality and low-quality images. The proposed method can detect forgeries from very challenging images. Figure~\ref{fig:better_results} displays some examples that are falsely classified by XceptionNet but correctly by our method.

\begin{table*}[!t]
	\begin{center}
		\fontsize{7pt}{12pt}\selectfont
		\caption{Ablation Study on the Forensics++-Deepfakes dataset. (c0: RAW, c23: HQ, c40: LQ) }
		\begin{tabular}{l|c|c|c|c|c|c|c|c|c}
			\toprule
			No. & Network                  & Losses or Modules      & Training data & Test data & ACC & AUC & TAR$_{0.1\%}$ & TAR$_{0.01\%}$ & PBCA \\
			\midrule
			1 & Single XceptionNet    		& Cross Entropy       & c40 & c40  & 94.36 & 97.90 & 94.74 & 79.32 & $-$ \\
			2 & Single XceptionNet     		& Metric Loss       & c40 & c40 & 94.33 & 97.70 & 95.87 & 84.71 & $-$ \\
			3 & Single XceptionNet     		& Metric Loss+Attention  & c40 & c40 & 94.46 & 97.86 & 96.18 & 85.67 & 93.33\\
			4 & Single XceptionNet    		& Metric Loss+Attention  & c23 & c23 & 98.93 & 99.85 & 99.82 & 98.89 & 96.34\\
			\midrule
			5 & 2-Branch XceptionNet             & Metric Loss        & (c23,c40) & c40            & 94.71 & 97.91 & 96.69 & 86.67 & $-$\\
			6 & 2-Branch XceptionNet        		& Metric Loss+GAN       & (c23,c40) & c40            & 94.85 & 98.47 & 96.71 & 84.27 & $-$\\
			7 & 2-Branch XceptionNet       		& Metric Loss+Attention     & (c23,c40) & c40            & 95.04 & 98.05 & 97.16 & 86.53 & 93.51\\
			8 & 2-Branch XceptionNet       		& Metric Loss+Att.Tran.     & (c23,c40) & c40            & 95.16 & 98.18 & 97.28 & 86.78 & 93.64\\
			9 & 2-Branch XceptionNet        		& Metric Loss+Att.Tran.+GAN     & (c23,c40) & c40             & 95.30 & 98.76 & 97.64 & 84.25 & 93.77\\
			10 & 2-Branch XceptionNet        		& Metric Loss+Att.Tran.+GAN     & (c0,c40) & c40             & 95.21 & 98.25 & 96.93 & 81.96 & 93.65\\
			11 & 2-Branch XceptionNet        		& Metric Loss+Att.Tran.+GAN     & (c0,c40)(c23,c40) & c40    & 95.38 & 98.68 & 97.59 & 83.85 & 93.86\\
			\bottomrule
		\end{tabular}
		\label{tab:ablation_methods}
	\end{center}
	\vspace{-4mm}
\end{table*}

\subsection{Ablation Study}
We do ablation experiments on both single XceptionNet and our proposed two-branch XceptionNet to explore benefits of metric loss, attention transfer and GAN. The tuple (c23, c40) represents that the high-quality branch takes in c23 compression-level images and the low-quality branch takes in c40 compression-level images.

\textbf{Benefit from GAN}:
As shown in Table~\ref{tab:ablation_methods}, the-two branch network trained with GAN obtains much higher ACC and AUC than the network without GAN, holding an improvement of $0.14\%$ on ACC and  $0.58\%$ on AUC. What's more, GAN also brings an improvement of $0.13\%$ on PBCA~, as shown in the comparison of row ~8 and row~9 in Table~\ref{tab:ablation_methods}.
Since GAN aligns the distribution of features from two branches, it will be more easier for the following network to learn a common representation.

\textbf{Benefit from Attention Transfer}:
We firstly explore the effect of attention layer in the single XceptionNet. Results in Table~\ref{tab:ablation_methods} shows that the attention layer can improve all evaluation metrics to some extent, for it enforces the network concentrating on the tampered region. Attention layer will improve the ACC of $0.13\%$ and AUC of $0.16\%$ in the single XceptionNet, as shown in row 2 and row 3 in Table~\ref{tab:ablation_methods}.
Two-branch network with attention transfer gets better classification accuracy  than the two-branch network without attention layer, since the low-quality branch gets more accurate location. There is $0.45\%$, $0.27\%$, $0.59\%$, $0.11\%$ performance improvement on ACC, AUC, TAR$_{0.1\%}$ and TAR$_{0.01\%}$, as can be seen in row 5 and row 8 in Table~\ref{tab:ablation_methods}. We further verify that the attention transfer is more effective than only using attention layer without transfer, comparing row 7 and row 8 in Table~\ref{tab:ablation_methods}. As shown in Fig.~\ref{fig:mask}, we plot attention maps of the high-quality and low-quality branch on the four manipulation-specific dataset of FaceForensics++.

\textbf{Benefit from Metric Loss}:
To reveal the effects of metric loss, we train the single XceptionNet supervised by cross-entropy loss and metric loss respectively. As shown in Table \ref{tab:ablation_methods}, the single model trained with metric loss has similar AUC and accuracy compared with the single model trained with cross entropy, but achieves much higher TAR$_{0.1\%}$ of  $1.13\%$  and TAR$_{0.01\%}$ of  $5.4\%$ than cross-entropy loss, comparing row 1 and row 2 in Table~\ref{tab:ablation_methods}. It proves that metric loss performs better at lower FAR.

\begin{figure*} [!t]
\centering
 \subfigure[Histogram of distances] {
  \label{fig:a}
  \includegraphics[width=0.836\columnwidth]{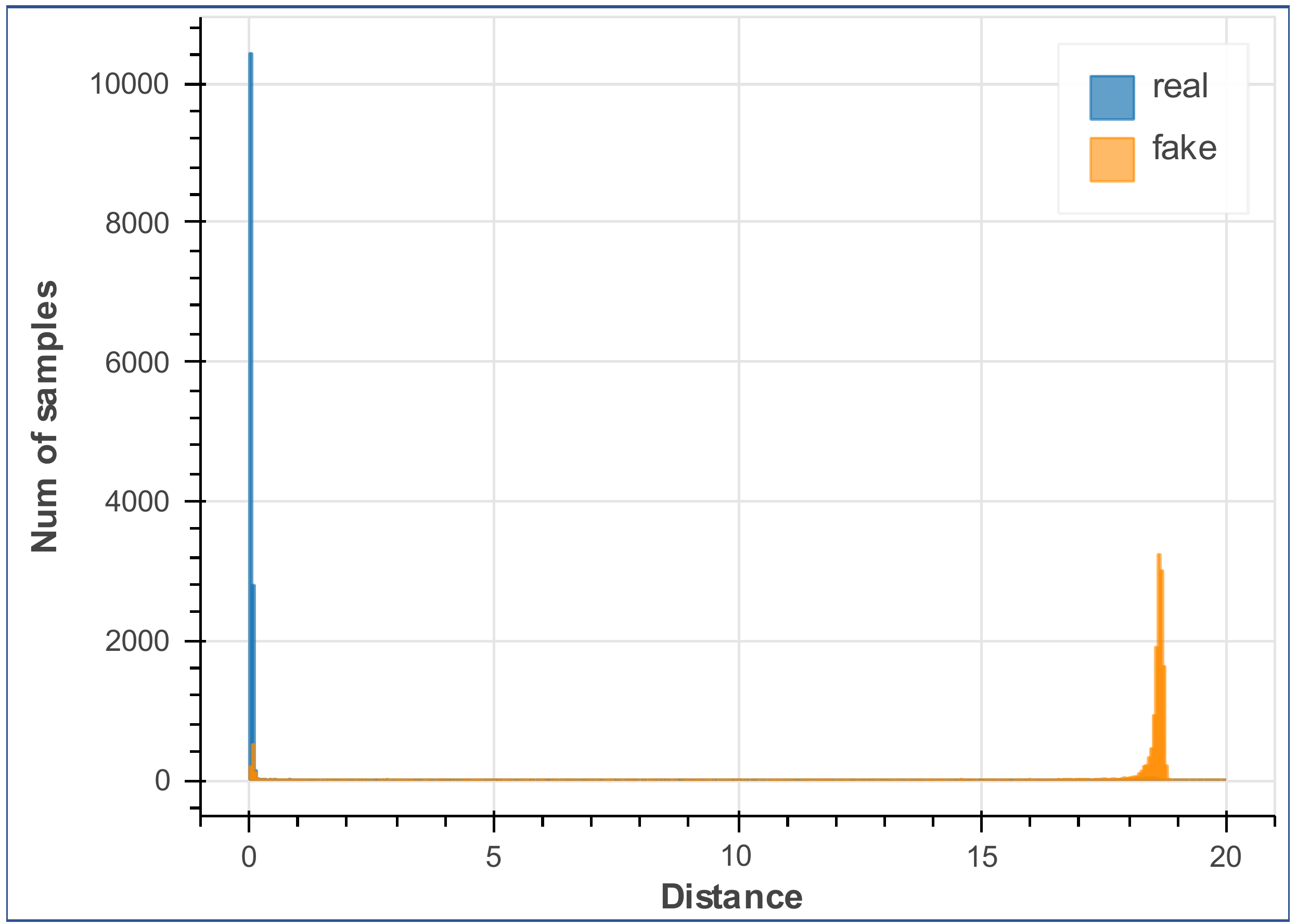}
 } \hspace{6mm}
 \subfigure[Visualization of features using t-sne~\cite{56-van2008visualizing}] {
  \label{fig:b}
  \includegraphics[width=0.91\columnwidth]{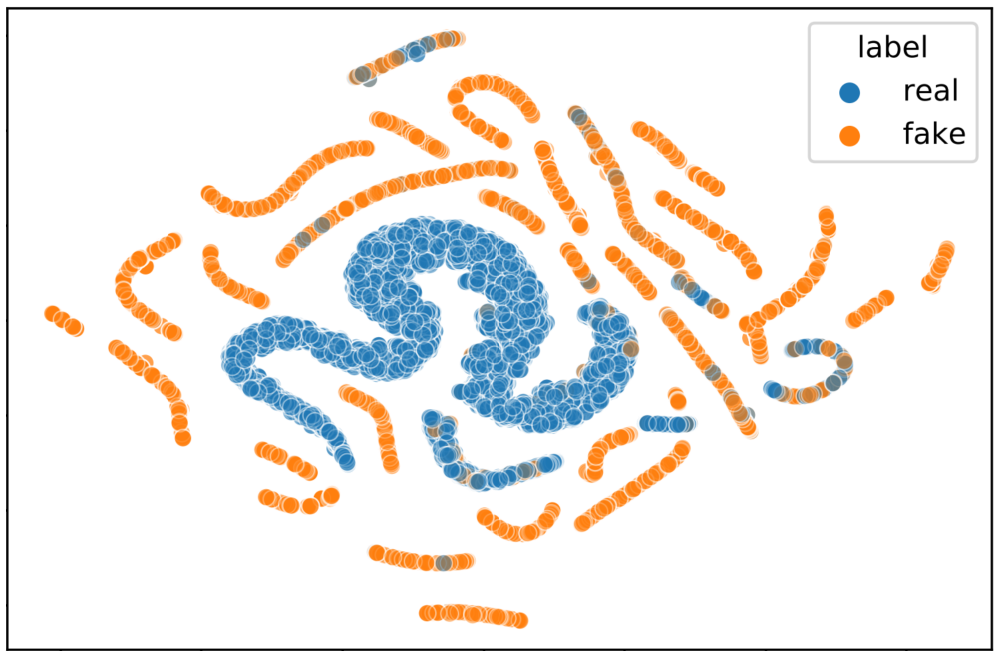}
 }
 \caption{Feature distribution of images from FaceForensics++(LQ)}
 \label{fig:dis}
\end{figure*}

It is worth noting that the metric loss of the two-branch network has one more component that reduces embedding distance of paired images than the single network. The two-branch network supervised by metric loss achieves higher performance on all evaluation metrics than single model. The two-branch network achieve an improvement of about $0.4\%$ on ACC, $0.2\%$ on AUC, $0.9\%$ on TAR$_{0.1\%}$ and $2\%$ on TAR$_{0.01\%}$, as shown row 2 and row 5 in Table~\ref{tab:ablation_methods}. It is because that, the metric loss requires the low-quality branch learning a similar feature representation with high-quality branch, enforcing the low-quality branch extracting the rare meaningful feature as much as possible. As shown in Fig.~\ref{fig:dis}, we plot histogram of the distance from the origin and use t-sne~\cite{56-van2008visualizing} to visualize distribution of feature. We feed paired HQ(c23) and LQ(c40) test data to the high-quality and low-quality branch respectively to obtain the distances. In Fig.~\ref{fig:dis}, we can see that real and fake data are separated with a large margin.

We can also find that, the model trained on (c23, c40) pairs performs better than (c0, c40) pairs, when comparing row 9 and row 10 in Table~\ref{tab:ablation_methods}. The two-branch network trained with attention transfer, metric loss and GAN on (c23, c40) pairs achieves the best result in ACC of  $95.30\%$, AUC of  $98.76\%$, and TAR$_{0.1\%}$ of $97.64\%$, and the second best in PBCA of $93.77\%$, as shown in row 9 of Table~\ref{tab:ablation_methods}. The reason may be that the distribution difference of c23-data and c40-data is smaller, which makes the network learn a common feature space more easily. But there is still a large gap on all evaluation metrics  compared with the single XceptionNet directly trained and test on c23-data, by comparing row 4 and row 9 in Table~\ref{tab:ablation_methods}, for lots of determinative features are lost after compression.

\section{Conclusion}
In this paper, we studied the detection of the compressed facial forgery images. The proposed two-branch network performed an accurate and stable facial forgery detection with the help of an adversarial learning strategy, a metric loss, and an attention transfer. The adversarial learning with the input of compressed and uncompressed pairs drove the network to extract compression-insensitive features. The metric loss maximized the distance of genuine and forgery data. The attention transfer improved the prediction of region of forgery manipulation. To the best of our knowledge, this was the first work that uses paired images across different compression levels to improve the capacity of handling compressed facial forgeries. Experiments showed that, the proposed method achieved the state-of-art performance on the benchmark datasets.

{\small
\bibliographystyle{ieee_fullname}
\bibliography{reference}
}

\end{document}